\documentclass[runningheads]{llncs}
\usepackage[T1]{fontenc}
\usepackage{graphicx}
\usepackage{listings}
\usepackage{algorithm}
\usepackage{algcompatible}
\usepackage{algpseudocode}

\usepackage{amsmath,amssymb}
\usepackage{cleveref}
\usepackage{subcaption}
\usepackage{stmaryrd}
\usepackage{booktabs}
\usepackage{tabularx}
\usepackage{dirtytalk}
\usepackage{multirow}
\usepackage{adjustbox}
\usepackage{bm}

\usepackage[acronym]{glossaries}
\newacronym{us}{US}{Ultrasound}
\newacronym{fps}{fps}{frames per second}
\newacronym{ef}{LVEF}{Left Ventricular Ejection Fraction}
\newacronym{es}{ES}{End-Systolic}
\newacronym{ed}{ED}{End-Diastolic}
\newacronym{sv}{SV}{Systolic Volume}
\newacronym{cdm}{CDM}{Cascaded Diffusion Model}
\newacronym{edm}{EDM}{Elucidated Diffusion Model}
\newacronym{ml}{ML}{Machine Learning}
\newacronym{mri}{MRI}{Magnetic Resonance Imaging}
\newacronym{ct}{CT}{Computed Tomography}

\newcommand{\ua}{\uparrow}
\newcommand{\da}{\downarrow}

\makeatletter
\newcommand*{\inlineequation}[2][]{%
  \begingroup
    \refstepcounter{equation}%
    \ifx\\#1\\%
    \else
      \label{#1}%
    \fi
    \relpenalty=10000 %
    \binoppenalty=10000 %
    \ensuremath{%
      #2%
    }%
    ~\@eqnnum
  \endgroup
}
\makeatother

\begin{document}
\title{Feature-Conditioned Cascaded Video Diffusion Models for Precise Echocardiogram Synthesis}
\titlerunning{Feature-Contioned Echocardiogram Synthesis}
%
\author{
Hadrien~Reynaud\inst{1,2} \and 
Mengyun~Qiao\inst{2,3}
Mischa~Dombrowski\inst{4} \and 
Thomas~Day\inst{5,6} \and
Reza~Razavi\inst{5,6} \and
Alberto~Gomez\inst{5,7} \and
Paul~Leeson\inst{7,8} \and 
Bernhard~Kainz\inst{2,4}
}

\institute{
UKRI CDT in AI for Healthcare, Imperial College London, London, UK \email{hadrien.reynaud19@imperial.ac.uk} \and
Department of Computing, Imperial College London, London, UK \and
Department of Brain Sciences and Data Science Institute, Imperial College London, London, UK \and
Friedrich--Alexander University Erlangen--N\"urnberg, DE \and
School of Biomedical Engineering and Imaging Sciences, King's College London, London, UK \and
Department of Congenital Cardiology, Evelina London Children's Healthcare, Guy's and St Thomas' NHS Foundation Trust, London, UK \and
Ultromics Ltd, Oxford, UK \and 
John Radcliffe Hospital, Cardiovascular Clinical Research Facility, Oxford, UK
}

\authorrunning{H. Reynaud et al.}

\maketitle              

\begin{abstract}

Image synthesis is expected to provide value for the translation of machine learning methods into clinical practice. Fundamental problems like model robustness, domain transfer, causal modelling, and operator training become approachable through synthetic data. Especially, heavily operator-dependant modalities like Ultrasound imaging require robust frameworks for image and video generation. So far, video generation has only been possible by providing input data that is as rich as the output data, \emph{e.g.}, image sequence plus conditioning in~$\rightarrow$~video out. However, clinical documentation is usually scarce and only single images are reported and stored, thus retrospective patient-specific analysis or the generation of rich training data becomes impossible with current approaches. 
In this paper, we extend elucidated diffusion models for video modelling to generate plausible video sequences from single images and arbitrary conditioning with clinical parameters. 
We explore this idea within the context of echocardiograms by looking into the variation of the Left Ventricle Ejection Fraction, the most essential clinical metric gained from these examinations. We use the publicly available EchoNet-Dynamic dataset for all our experiments.
Our image to sequence approach achieves an $R^2$ score of 93\%, which is 38 points higher than recently proposed sequence to sequence generation methods. Code and models will be available at: \url{https://github.com/HReynaud/EchoDiffusion}.

\end{abstract}
\section{Introduction}
\gls{us} is widely used in clinical practice because of its availability, real-time imaging capabilities, lack of side effects for the patient and flexibility. \gls{us} is a dynamic modality that heavily relies on operator experience and on-the-fly interpretation, which requires many years of training and/or \gls{ml} support that can handle image sequences. 
However, clinical reporting is conventionally done via single, selected images that rarely suffice for clinical audit or as training data for \gls{ml}. 
Simulating \gls{us} from anatomical information, \emph{e.g.} \gls{ct} \cite{shams_real-time_2008}, \gls{mri} \cite{salehi_patient-specific_2015} or computational phantoms \cite{jensen_simulation_2004,segars_4d_2010}, has been considered as a possible avenue to provide more \gls{us} data for both operator and \gls{ml} training. However, simulations are usually very computationally expensive due to complex scattering, reflection and refraction of sound waves at tissue boundaries during image generation. 
Therefore, the image quality of \gls{us} simulations has not yet met the necessary quality to support tasks such as cross-modality registration, multi-modal learning, and robust decision support for image analysis during \gls{us} examinations. More recently, generative deep learning methods have been proposed to address this issue. While early approaches show promising results, they either focus on generating individual images \cite{liang_sketch_2022} or require video input data and further conditioning to provide useful results \cite{liang_weakly-supervised_2022,reynaud_dartagnan_2022}. Research in the field of image-conditioned video generation is very scarce \cite{song_talking_2019} and, to the best of our knowledge, we are the first to apply it to medical imaging.

\noindent\textbf{Contribution: } In this paper, we propose a new method for video diffusion \cite{ho_imagen_2022,singer_make--video_2022} based on the \gls{edm} \cite{karras_elucidating_2022} that allows to synthesise plausible video data from single frames together with precise conditioning on interpretable clinical parameters, \emph{e.g.}, \gls{ef} in echocardiography. 
This is the first time diffusion models have been extended for \gls{us} image and video synthesis. Our contributions are three-fold:
(1) We show that discarding the conventional text-embeddings \cite{ho_imagen_2022,ramesh_zero-shot_2021,rombach_high-resolution_2022,saharia_photorealistic_2022,singer_make--video_2022} to control the reverse diffusion process is desirable for medical use cases where very specific elements must be precisely controlled; 
(2) We quantitatively improve upon existing methods \cite{reynaud_dartagnan_2022} for counterfactual modelling, \emph{e.g.}, when doctors try to answer questions like ``how would the scan of this patient look like if we would change a given clinical parameter?'';  
(3) We show that fine-grained control of the conditioning leads to precise data generation with specific properties and outperforms the state-of-the-art when using such data, for example, for the estimation of \gls{ef} in patients that are not commonly represented in training databases. 

\noindent\textbf{Related work}:
\noindent\textbf{Video Generation} has been a research area within computer vision for many years now. Prior works can be organized in three categories: (1) pixel-level autoregressive models \cite{babaeizadeh_fitvid_2021,finn_unsupervised_2016,kalchbrenner_video_2017}, (2) latent-level autoregressive model coupled with generators or up-samplers \cite{babaeizadeh_stochastic_2018,kumar_videoflow_2020} and (3) latent-variable transformer-based models with up-samplers \cite{gupta_maskvit_2022,villegas_phenaki_2022}. Diffusion models have recently shown reasonable performance on low temporal and spatial resolutions \cite{ho_video_2022} as well as on longer samples with high definition image quality \cite{ho_imagen_2022,singer_make--video_2022} conditioned on text inputs. Recently, \cite{yang_diffusion_2022} combined an autoregressive pixel-level model with a diffusion-based pipeline that predicts a correction of the frame, while \cite{esser_structure_2023} presents an autoregressive latent diffusion model. 

\noindent\textbf{Ultrasound simulation} has been attempted with three major approaches: (1) physics-based simulators \cite{jensen_simulation_2004,shams_real-time_2008}, (2) cross-modality registration-based methods \cite{ledesma-carbayo_spatio-temporal_2005} and (3) deep-learning based methods, usually conditioned on \gls{us}, \gls{mri} or \gls{ct} image priors~\cite{salehi_patient-specific_2015,teng_interactive_2020,tomar_content-preserving_2021} to condition the anatomy of the generated \gls{us} images. 
Cine-ultrasound has also attracted some interest. \cite{liang_weakly-supervised_2022} presents a motion-transfer-based method for pelvic \gls{us} video generation, while \cite{reynaud_dartagnan_2022} proposes a causal model for generating echocardiograms conditioned on arbitrary \gls{ef}. 

\noindent\textbf{\gls{ef}}\label{sec:reg_ef}
is a major metric in the assessment of cardiac function and diagnosis of cardiomyopathy. The EchoNet-dynamic dataset \cite{ouyang_video-based_2020} is used as the go-to benchmark for \gls{ef}-regression methods. Various works \cite{mokhtari_echognn_2022,reynaud_ultrasound_2021} have attempted to improve on \cite{ouyang_video-based_2020} but the most reproducible method remains the use of an R2+1D model trained over fixed-length videos. The \texttt{R2+1D\_18} trained for this work achieves an $R^2$ score of 0.81 on samples of 64 frames spanning 2 seconds.
\section{Method}

\begin{figure}
    \centering
    \vspace{-0.5cm}
    \includegraphics[trim={0 0.35cm 0 0},clip,width=\linewidth]{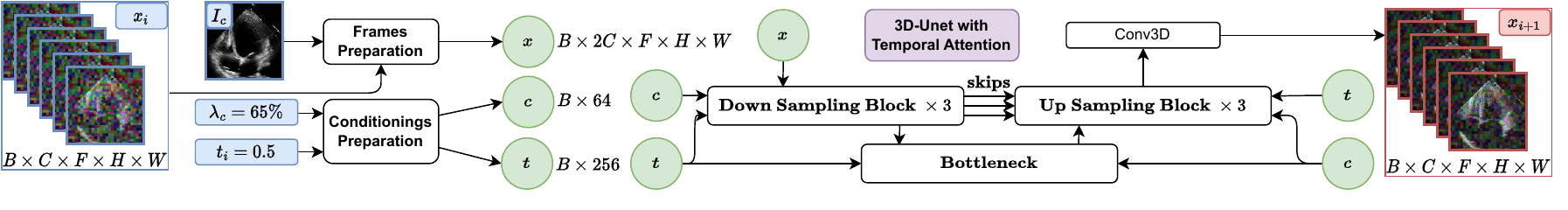}
    \caption{Summarized view of our Model. 
    Inputs (blue): a noised sample $x_i$, a diffusion step $t_i$, one anatomy image $I_c$, and one LVEF $\lambda_c$. 
    Output (red): a slightly denoised version of $x_i$ named $x_{i+1}$. 
    See Appendix Fig.~1 for more details.}
    \label{fig:narrow_model}
    \vspace{-0.5cm}
\end{figure}
Diffusion probabilistic models \cite{ho_denoising_2020,sohl-dickstein_deep_2015,song_denoising_2022} are the most recent family of generative models. In this work, we follow the definition of the \gls{edm} from \cite{karras_elucidating_2022}. 
Let $q(\bm{x})$ represent the real distribution of our data, with a standard deviation of $\sigma_q$. A family of distributions $p(\bm{x}; \sigma)$ can be obtained by adding i.i.d Gaussian noise with a standard deviation of $\sigma$ to the data. When $\sigma_{\text{max}} \gg \sigma_q$, the distribution $p(\bm{x}; \sigma_{\text{max}})$ is essentially the same as pure Gaussian noise. The core idea of diffusion models is to sample a pure noise data point $\bm{x}_0 \sim \mathcal{N}(\bm{0}, \sigma_{\text{max}}^{2}\textbf{I})$ and then progressively remove the noise, generating images $\bm{x}_i$ with standard deviation $\sigma_i$ such that
$\sigma_{\text{max}} = \sigma_0 > \sigma_1 > ... > \sigma_N = 0 $, and $\bm{x}_i \sim p(\bm{x}; \sigma_i)$. The final image $\bm{x}_N$ produced by this process is thus distributed according to $q(\bm{x})$, the true distribution of the data. 
To perform the reverse diffusion process, we define a denoising function $D(\bm{x}, \sigma)$ trained to minimize the $L_2$ denoising error for all samples drawn from $q$ for every $\sigma$ such that: 
\inlineequation[eq:loss]{\mathcal{L}~=~\mathbb{E}_{\bm{y}\sim q}\mathbb{E}_{\bm{n} \sim \mathcal{N}(\bm{0}, \sigma^{2}\textbf{I})} ||D(\bm{y}+\bm{n};\sigma) - \bm{y}||^2_2}
where $\bm{y}$ is a training data point and $\bm{n}$ is noise. 
By following the definition of ordinary differential equations (ODE) we can continuously increase or decrease the noise level of our data point by moving it forward or backward in the diffusion process, respectively. To define the ODE we need a schedule $\sigma(t)$ that sets the noise level given the time step $t$, which we set to $\sigma(t) = t$ . The probability flow ODE's characteristic property is that moving a sample $\bm{x}_a \sim p(\bm{x}_a;\sigma(t_a)) $ from the diffusion step $t_a$ to $t_b$ with $t_a > t_b$ or $t_a < t_b$ should result in a sample $\bm{x}_b \sim p(\bm{x}_b;\sigma(t_b)) $ and this requirement is satisfied by setting $\text{d}\bm{x} = -\dot{\sigma}(t)\sigma(t)\nabla_x\text{log}~p(\bm{x};\sigma(t))\text{d}t $ where $\dot{\sigma}$ denotes the time derivative and $\nabla_x\text{log}~p(\bm{x};\sigma)$ is the score function. 
From the score function, we can thus write $\nabla_x\text{log}~p(\bm{x};\sigma) = (D(\bm{x};\sigma) - \bm{x})/\sigma^2$ in the case of our denoising function, such that the score function isolates the noise from the signal $\bm{x}$ and can either amplify it or diminish it depending on the direction we take in the diffusion process. 
We define $D(\bm{x};\sigma)$ to transform a neural network $F$, which can be trained inside $D$ by following the loss described in \Cref{eq:loss}. The \gls{edm} also defines a list of four important preconditionings which are defined as 
$c_{\text{skip}}(\sigma) = (\sigma_q^2)/(\sigma^2+\sigma_q^2)$, $c_{\text{out}}(\sigma) = \sigma * \sigma_q * 1/(\sigma_q^2 * \sigma^2)^{0.5}$, $c_{\text{in}}(\sigma) = 1/(\sigma_q^2 * \sigma^2)^{0.5}$ and  $c_{\text{noise}}(\sigma) = \text{log}(\sigma_t)/4$
where $\sigma_q$ is the standard deviation of the real data distribution.
In this paper, we focus on generating temporally coherent and realistic-looking echocardiograms.
We start by generating a low resolution, low-frame rate video $\bm{v}_0$ from noise and condition on arbitrary clinical parameters and an anatomy instead of the commonly used text-prompt embeddings~\cite{ho_imagen_2022,singer_make--video_2022}.
Then, the video is used as conditioning for the following diffusion model, which generates a temporally and/or spatially upsampled video $\bm{v}_1$ resembling $\bm{v}_0$, following the \gls{cdm} \cite{ho_cascaded_2022} idea. Compared to image diffusion models, the major change to the Unet-based architecture is to add time-aware layers, through attention, at various levels as well as 3D convolutions (see Fig.~\ref{fig:narrow_model} and Appendix Fig.~1). 
For the purpose of this research, we extend \cite{ho_imagen_2022} to handle our own set of conditioning inputs, which are a single image $\bm{I}_c$ and a scalar value $\lambda_c$, while following the \gls{edm} setup, which we apply to video generation.
We formally define the denoising models in the cascade as $D_{\theta_s}$ where $s$ defines the rank (stage) of the model in the cascade, and where $D_{\theta_0}$ is the base model. The base model is defined as: 
\begin{align*}
    D_{\theta_0}(\bm{x};\sigma,\bm{I}_c,\lambda_c) = c_{\text{skip}}(\sigma)\bm{x} + c_{\text{out}}(\sigma) F_{\theta_0}(c_{\text{in}}(\sigma)\bm{x};c_{\text{noise}}(\sigma), \bm{I}_c,\lambda_c)), 
\end{align*}
where $F_{\theta_0}$ is the neural network transformed by $D_{\theta_0}$ and $D_{\theta_0}$ outputs $\bm{v}_0$.
For all subsequent models in the cascade, the conditioning remains similar, but the models also receive the output from the preceding model, such that:
\begin{align*}
    D_{\theta_s}(\bm{x};\sigma,\bm{I}_c,\lambda_c, \bm{v}_{s-1}) = c_{\text{skip}}(\sigma)\bm{x} + c_{\text{out}}(\sigma) F_{\theta_s}(c_{\text{in}}(\sigma)\bm{x};c_{\text{noise}}(\sigma), \bm{I}_c,\lambda_c, \bm{v}_{s-1})).
\end{align*}
This holds $\forall{s} > 0$ and inputs $\bm{I}_c, \bm{v}_{s-1}$ are rescaled to the spatial and temporal resolutions expected by the neural network $F_{\theta_s}$ as a pre-processing. We apply the robustness trick from \cite{ho_cascaded_2022}, \emph{i.e}, we add a small amount of noise to real videos $\bm{v}_{s-1}$ during training, when using them as conditioning, in order to mitigate domain gaps with the generated samples $\bm{v}_{s-1}$ during inference.

Sampling from the \gls{edm} is done through a stochastic sampling method. We start by sampling a noise sample $\bm{x}_0 \sim \mathcal{N}(\bm{0}, t^2_0\textbf{I})$, where $t$ comes from our previously defined $\sigma(t_i)=t_i$ and sets the noise level. 
We follow \cite{karras_elucidating_2022} and set constants
$S_{\text{noise}}=1.003, S_{t_\text{min}}=0.05, S_{t_\text{max}}=50$ and one constant $S_{\text{churn}}$ dependent on the model. These are used to compute $\gamma_i(t_i) = \min(S_{\text{churn}}/N, \sqrt{2}-1)$ $\forall t_i \in [S_{t_\text{min}}, S_{t_\text{max}}]$ and 0 otherwise, where $N$ is the number of sampling steps. Then $\forall i \in \{0,...,N-1\}$, we sample $\bm{\epsilon}_i \sim \mathcal{N}(\bm{0}, S_{\text{noise}}\textbf{I})$
and compute a slightly increased noise level $\hat{t}_i = (\gamma_i(t_i)+1)t_i$, 
which is added to the previous sample $\hat{\bm{x}}_i = \bm{x}_i + (\hat{t}^2_i - t^2_i)^{0.5}\bm{\epsilon}_i$. 
We then execute the denoising model $D_{\theta}$ on that sample and compute the local slope 
$\bm{d}_i = (\hat{\bm{x}}_i - D_{\theta}(\hat{\bm{x}}_i;\hat{t}_i))/\hat{t}_i $ 
which is used to predict the next sample 
$ \bm{x}_{i+1}~=~\hat{\bm{x}}_i~+~(\hat{t}_{i+1}~-~\hat{t}_i)\bm{d}_i $. 
At every step but the last (\emph{i.e:} $\forall i \neq N-1$), we apply a correction to $\bm{x}_{i+1}$ such that: 
$\bm{d}_i' = (\bm{x}_{i+1} - D_{\theta}(\bm{x}_{i+1};t_{i+1}))/t_{i+1} $ and 
$\bm{x}_{i+1} = \hat{\bm{x}}_i + (t_{i+1} - \hat{t}_i) (\bm{d}_i + \bm{d}_i') / 2$. The correction step doubles the number of executions of the model, and thus the sampling time per step, compared to DDPM\cite{ho_denoising_2020} or DDIM\cite{song_denoising_2022}.
The whole sampling process is repeated sequentially for all models in the cascaded \gls{edm}. Models are conditioned on the previous output video $\bm{v}_{s-1}$ inputted at each step of the sampling process, with the frame conditioning $\bm{I}_c$ as well as the scalar value $\lambda_c$. 

\noindent\textbf{Conditioning}: Our diffusion models are conditioned on two components. First, an \textit{anatomy}, which is represented by a randomly sampled frame $I_c$. It defines the patient's anatomy, but also all the information regarding the visual style and quality of the target video. These parameters cannot be explicitly disentangled, and we therefore limit ourselves to this approach. Second, we condition the model on clinical parameters $\lambda_c$. This is done by discarding the text-encoders that are used in \cite{ho_video_2022,singer_make--video_2022} and directly inputting normalized clinical parameters into the conditional inputs of the Unets. By doing so, we give the model fine-grained control over the generated videos, which we evaluate using task-specific metrics.

\noindent\textbf{Parameters}: As video diffusion models are still in their early stage, there is no consensus on which are the best methods to train them. In our case, we define, depending on our experiment, 1-, 2- or 4-stages \gls{cdm}s. We also experiment with various schedulers and parametrizations of the model. \cite{salimans_progressive_2022,song_denoising_2022} show relatively fast sampling techniques which work fine for image sampling. However, in the case of video, we reach larger sampling times as we sample 64 frames at once. 
We therefore settled for the \gls{edm} \cite{karras_elucidating_2022}, which presents a method to sample from the model in much fewer steps, largely reducing sampling times. We do not observe any particular speed-up in training and would argue, from our experience, that the v-parametrization \cite{song_denoising_2022} converges faster. We experimentally find our models to behave well with parameters close to those suggested in~\cite{karras_elucidating_2022}.
\section{Experiments}
\noindent\textbf{Data: } 
We use the EchoNet-Dynamic~\cite{ouyang_video-based_2020} dataset, a publicly available dataset that consists of 10,030 4-chamber cardiac ultrasound sequences, with a spatial resolution of $112 \times 112$ pixels. Videos range from 0.7 to 20.0 seconds long, with frame rates between 18 and 138 \gls{fps}. Each video has 3 channels, although most of them are greyscale. 
We keep the original data split of EchoNet-Dynamic which has 7465 training, 1288 validation and 1277 testing videos.We only train on the training data, and validate on the validation data.
In terms of labels, each video comes with an \gls{ef} score $\lambda \in [0,100]$, estimated by a trained clinician. 
At every step of our training process, we pull a batch of videos, which are resampled to 32 \gls{fps}. For each video, we retrieve its corresponding ground truth \gls{ef} as well as a random frame. After that, the video is truncated or padded to 64 frames, in order to last 2 seconds, which is enough to cover any human heartbeat. 
The randomly sampled frame is sampled from the same original video as the 64-frames sample, but may not be contained in those 64 frames, as it may come from before or after that sub-sample.

\noindent\textbf{Architectural variants: } 
We define three sets of models, and present them in details in Table 1 of the Appendix. We call the models \textit{X}-Stage Cascaded Models (\textit{X}SCM) and present the models' parameters at every stage. Every \gls{cdm} starts with a \textit{Base} diffusion model that is conditioned on the \gls{ef} and one conditional frame. The subsequent models perform either temporal super resolution (TSR), spatial super resolution (SSR) or temporal and spatial super resolution (TSSR). TSR, SSR and TSSR models receive the same conditioning inputs as the Base model, along with the output of the previous-stage model. Note that \cite{ho_imagen_2022} does not mention TSSR models and \cite{singer_make--video_2022} states that extending an SSR model to perform simultaneous temporal and spatial up-sampling is too challenging.

\noindent\textbf{Training: }
All our models are trained from scratch on individual cluster nodes, each with 8 $\times$ NVIDIA A100. We use a per-GPU batch size of 4 to 8, resulting in batches of 32 to 64 elements after gradient accumulation. 
The distributed training is handled by the \texttt{accelerate} library from HuggingFace. We did not see any speed-up or memory usage reduction when enabling mixed precision and thus used full precision. As pointed out by \cite{ho_cascaded_2022} all models in a \gls{cdm} can be trained in parallel which significantly speeds up experimentation. We empirically find that training with a learning rate up to $5*10^{-4}$ is stable and reaches good image quality. We use an Exponential Moving Average (EMA) copy of our model to smooth out the training. We train all our models' stages for 48h, \emph{i.e.}, the 2SCM and 4SCM \gls{cdm}s are proportionally more costly to train than the 1SCM. As noted by \cite{ho_imagen_2022,singer_make--video_2022} training on images and videos improves the overall image quality. As our dataset only consists of videos, we simply deactivate the time attention layers in the Unet with a 25\% chance during training, for all models.

\noindent\textbf{Results: }
We evaluate our models' video synthesis capabilities on two objectives: \gls{ef} accuracy ($R^2$, MAE, RMSE) and image quality (SSIM, LPIPS, FID, FVD). We formulate the task as counterfactual modelling, where we set 
(1) a random conditioning frame as confounder, 
(2) the ground-truth \gls{ef} as a factual conditioning, and 
(3) a random \gls{ef} in the physiologically plausible range from 15\% to 85\% as counterfactual conditioning.
For each ground truth video, we sample three random starting noise samples and conditioning frames. We use the \gls{ef} regression model to create a feedback loop, following what \cite{reynaud_dartagnan_2022} did, even though their model was run $100\times$ per sample instead of $3\times$. For each ground truth video, we keep the sample with the best \gls{ef} accuracy to compute all our scores over 1288 videos for each model.
\begin{table}
    \centering
    \begin{tabularx}{\textwidth}{lllccXXcXcXc}
        \toprule
        Model & Task & Res. & Frames       & S. time       & $R^2\ua$ & MAE$\da$ & RMSE$\da$ & SSIM$\ua$ & LPIPS$\da$ & FID$\da$ & FVD$\da$ \\
        \midrule
        1SCM  & Gen. & 112  & 16$^{\ddag}$ & 62s$^{\ddag}$ & 0.64     & 9.65     & 12.2      & 0.53      & 0.21       & 12.3     & 60.5     \\
        2SCM  & Gen. & 112  & 64           & 146s          & 0.89     & 4.81     & 6.69      & 0.53      & 0.24       & 31.7     & 141      \\
        4SCM  & Gen. & 112  & 64           & 279s          & 0.93     & 3.77     & 5.26      & 0.48      & 0.25       & 24.6     & 230      \\
        \midrule
        1SCM  & Rec. & 112  & 16$^{\ddag}$ & 62s $^{\ddag}$& 0.76     & 4.51     & 6.07      & 0.53      & 0.21       & 13.6     & 89.7     \\
        2SCM  & Rec. & 112  & 64           & 146s          & 0.93     & 2.22     & 3.35      & 0.54      & 0.24       & 31.4     & 147      \\
        4SCM  & Rec. & 112  & 64           & 279s          & 0.90     & 2.42     & 3.87      & 0.48      & 0.25       & 24.0     & 228      \\
        \bottomrule
        
        \end{tabularx}
    \caption{\label{tab:all_models}Metrics for all \gls{cdm}s. The \emph{Gen.} task is the counterfactual generation comparable to~\cite{reynaud_dartagnan_2022}, the \emph{Rec.} task is the factual reconstruction task. \emph{Frames} is the number of frames generated by the model, always spanning 2 seconds. $^{\ddag}$Videos are temporally upsampled to 64 frames for metric computation. \emph{S. time} is the sampling time for one video on an RTX A5000. $R^2$, MAE and RMSE are computed between the conditional \gls{ef} $\lambda_c$ and the regressed \gls{ef} using the model described in \Cref{sec:reg_ef}. SSIM, LPIPS, FID and FVD are used to quantify the image quality. LPIPS is computed with VGG \cite{simonyan_very_2015}, FID\cite{heusel_gans_2018} and FVD\cite{unterthiner_fvd_2019} with I3D. FID and FVD are computed over padded frames of $128 \times 128$ pixels.}
\end{table}

The results in Table~\ref{tab:all_models} show that increasing the frame rate improves model fidelity to the given \gls{ef}, while adding more models to the cascade decreases image quality. This is due to a distribution gap between true low-resolution samples and sequentially generated samples during inference. This issue is partially addressed by adding noise to real low-resolution samples during training, but the 1SCM model with only one stage still achieves better image quality metrics. However, the 2SCM and 4SCM models perform equally well on \gls{ef} metrics and outperform the 1SCM model thanks to their higher temporal resolution that precisely captures key frames of the heartbeat. The TSSR model, used in the 2SCM, yields the best compromise between image quality, \gls{ef} accuracy, and sampling times, and is compared to previous literature.
\begin{figure}
    \center
    \includegraphics[width=\linewidth]{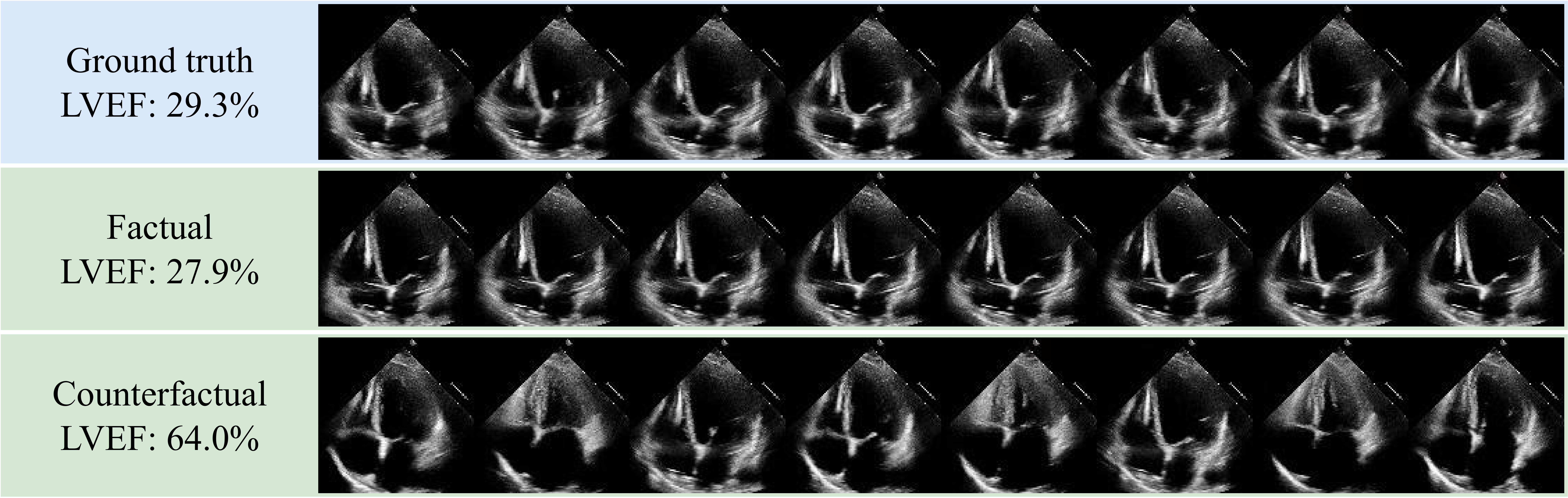}
    \caption{
    Top: Ground truth frames with 29.3\% \gls{ef}. 
    Middle: Generated factual frames, with estimated 27.9\% \gls{ef}.
    Bottom: Generated counterfactual frames, with estimated 64.0\% \gls{ef}.
    (Counter-)Factual frames are generated with the 1SCM, conditioned on the ground-truth anatomy.
    }
    \label{fig:us_example}
\end{figure}
\begin{table}
    \centering
    \begin{tabularx}{\textwidth}{llXXXXXX}
        \toprule
        Method                                   & Conditioning & Task   & S. time   & $R^2 \ua$ & MAE $\da$ & RMSE $\da$ & SSIM $\ua$ \\
        \midrule
        Dartagnan \cite{reynaud_dartagnan_2022}  & Video+EF     & Gen.   & $\sim$1s       & 0.51      & 15.7      & 18.4       & \textbf{0.79} \\
        2SCM                                     & Image+EF     & Gen.   & 146s      & \textbf{0.89}      & \textbf{4.81}      & \textbf{6.69}       & 0.53 \\
        \midrule
        Dartagnan \cite{reynaud_dartagnan_2022}  & Video+EF     & Rec.   & $\sim$1s       & 0.87      & 2.79      & 4.45       & \textbf{0.82}  \\
        2SCM                                     & Image+EF     & Rec.   & 146s      & \textbf{0.93}      & \textbf{2.22}      & \textbf{3.35}       & 0.54  \\
        \bottomrule
    \end{tabularx}
    \caption{\label{tab:comparison} Comparison of our 2SCM model to previous work. We try to reconstruct a ground truth video or to generate a new one. Our model is conditioned on a single frame and an \gls{ef}, while \cite{reynaud_dartagnan_2022} conditions on the entire video and an \gls{ef}. In both cases the LVEF is either the ground truth \gls{ef} (\emph{Rec.}) or a randomly sampled \gls{ef} (\emph{Gen.}).}
\end{table}

We outperform previous work for \gls{ef} regression: counterfactual video generation improves with our method by a large margin of 38 points for the $R^2$ score as shown in Table~\ref{tab:comparison}. The similarity between our factual and counterfactual results show that our time-agnostic confounding factor (\emph{i.e.} an image instead of a video) prevents entanglement, as opposed to the approach taken in \cite{reynaud_dartagnan_2022}. Our method does not score as high for SSIM as global image similarity metric, which is expected because of the stochasticity of the speckle noise. In \cite{reynaud_dartagnan_2022} this was mitigated by their data-rich confounder. Our results also match other video diffusion models \cite{esser_structure_2023,ho_imagen_2022,ho_video_2022,singer_make--video_2022} as structure is excellent, while texture tends to be more noisy as shown in Fig.~\ref{fig:us_example}.

\noindent\textbf{Qualitative study: }
We asked three trained clinicians (Consultant cardiologist $>$ 10 years experience, Senior trainee in cardiology $>$ 5 years experience, Chief cardiac physiologist $>$ 15 years experience) to classify 100 samples, each, as \emph{real} or \emph{fake}. Experts were not given feedback on their performance during the evaluation process and were not shown fake samples beforehand. All samples were generated with the 1SCM model or were true samples from the EchoNet-Dynamic dataset, resampled to 32fps and 2s. The samples were picked by alphabetical order from the validation set. Among the 300 samples evaluated, 130 (43.33\%) were real videos, 89 (29.67\%) were factual generated videos, and 81 (27.0\%) were counterfactual generated videos. The average expert accuracy was 54.33\%, with an inter-expert agreement of 50.0\%. More precisely, experts detected real samples with an accuracy of 63.85\%, 50.56\% for factual samples and 43.21\% for the counterfactual samples. The average time taken to evaluate each sample was 16.2s. 
We believe that these numbers show the video quality that our model reaches, and can put in perspective the SSIM scores from \Cref{tab:all_models}.

\noindent\textbf{Downstream task: } We train our \gls{ef} regression model on rebalanced datasets and resampled datasets. 
We rebalance the datasets by using our 4SCM model to generate samples for \gls{ef} values that have insufficient data. 
The resampled datasets are smaller datasets randomly sampled from the real training set. We show that, in small data regimes, using generated data to rebalance the dataset improves the overall performance. Training on 790 real data samples yields an $R^2$ score of 56\% while the rebalanced datasets with 790 samples, $\sim$50\% of which are real, reaches a better 59\% on a balanced validation set. This observation is mitigated when more data is available. See Appendix Table 2 for all our results.

\noindent\textbf{Discussion:}
Generating echocardiograms is a challenging task that differs from traditional computer vision due to the noisy nature of US images and videos. However, restricting the training domain simplifies certain aspects, such as not requiring a long series of \gls{cdm}s to reach the target resolution of $112 \times 112$ pixels and limiting samples to 2 seconds, which covers any human heartbeat. The limited pixel-intensity space of the data also allows for models with fewer parameters. In the future, we plan to explore other organs and views within the US domain, with different clinical conditionings and segmentation maps.

\section{Conclusion}
Our application of \gls{edm}s to \gls{us} video generation achieves state-of-the-art performance on a counterfactual generation task, a data augmentation task, and a qualitative study by experts. This significant advancement provides a valuable solution for downstream tasks that could benefit from representative foundation models for medical imaging and precise medical video generation.
\break
\break
\noindent\textbf{Acknowledgements:} 
This work was supported by Ultromics Ltd. and the UKRI Centre for Doctoral Training in Artificial Intelligence for Healthcare\break(EP/S023283/1).
The authors gratefully acknowledge the scientific support and HPC resources provided by the Erlangen National High Performance Computing Center (NHR@FAU) of the Friedrich-Alexander-Universität Erlangen-Nürnberg (FAU) under the NHR project b143dc PatRo-MRI. NHR funding is provided by federal and Bavarian state authorities. NHR@FAU hardware is partially funded by the German Research Foundation (DFG) – 440719683.

\newpage
\bibliographystyle{splncs04}
\bibliography{bibliography.bib}

\section*{Appendix}
\subsection*{Models architecture details}

\begin{table}[h!]
    \centering
    \begin{tabularx}{\textwidth}{lclcccccccX}
        \toprule
        Model & Stage & Mode & Res.            & Frames & FPS & Dims. & Layers & Steps &$S_{\text{churn}}$& Comments\\
        \midrule
        1SCM  & 1     & Base & $112\times 112$ & 16     & 8   & 64    & 2,2,2  & 64    & 160              & No bot. att. \\
        \midrule
        2SCM  & 1     & Base & $56\times 56$   & 16     & 8   & 64    & 2,2,2  & 32    &  80              &  \\
        2SCM  & 2     & TSSR & $112\times 112$ & 64     & 32  & 64    & 2,2,2  & 64    & 160              & Mem. Opti. \\
        \midrule
        4SCM  & 1     & Base & $56\times 56$  & 16      & 8   & 64    & 2,2,2  & 32    &  40              & - \\
        4SCM  & 2     & TSR  & $56\times 56$  & 32      & 16  & 64    & 2,2,2  & 32    &  80              & Mem. Opti. \\
        4SCM  & 3     & TSR  & $56\times 56$  & 64      & 32  & 64    & 2,2,2  & 32    & 160              & Mem. Opti. \\
        4SCM  & 4     & SSR  & $112\times112$ & 64      & 32  & 64    & 2,2,2  & 64    & 160              & No bot. att.\\
        \bottomrule
    \end{tabularx}
    \caption{\label{tab:scms} Overview of our \textit{X}SCM architectures. The resolution, frames and fps values are those of the outputs of each model, from which the input dimensions can be inferred. The Dims. (dimensions) values define the base number of channels in the model, which is doubled at each layer, while Layers states the number of ResNet blocks at each layer. The layer count defines the depth of the UNet model. The steps refer to the number of sampling steps. The comments indicate if we deactivated bottleneck attention (\textit{No bot. att.}) or used modified downsampling and upsampling blocks to reduce memory usage (\textit{Mem. Opti.}).}
    \centering
    \begin{tabularx}{0.65\textwidth}{Xcccccc}
        \toprule
        Name          & Bin Size & Fake \% & Total Size & $R^2$ & MAE  & RMSE \\
        \midrule
        Balanced      & 10       & 52.28\% & 790        & 0.59  & 8.73 & 11.19 \\
        Balanced      & 20       & 50.51\% & 1580       & 0.69  & 7.46 & 9.79 \\
        Balanced      & 50       & 50.18\% & 3950       & 0.75  & 6.61 & 8.72 \\
        \midrule
        Baseline      & -        & 0\%     & 790        & 0.56  & 8.62 & 11.62 \\
        Baseline      & -        & 0\%     & 1580       & 0.73  & 4.12 & 9.07 \\
        Baseline      & -        & 0\%     & 3950       & 0.77  & 6.32 & 8.46 \\
        \bottomrule
    \end{tabularx}
    \caption{Results for LVEF regression downstream task. Balanced models are trained on a rebalanced training set where we use our 4SCM to equalize the number of samples for each LVEF value with a 1\% step size, from 10\% to 89\%. The baseline models are trained by sampling randomly over the training data, following it's unbalanced distribution. We evaluate on 588 samples from a resampled dataset built from the validation and test sets where the LVEF are balanced as much as possible to be representative of the performance over the whole spectrum of plausible LVEF values.}
\end{table}

\newpage

\subsection*{Illustration of Model Architecture}

\begin{figure}[!h]
    \centering
    \includegraphics[width=0.55\linewidth]{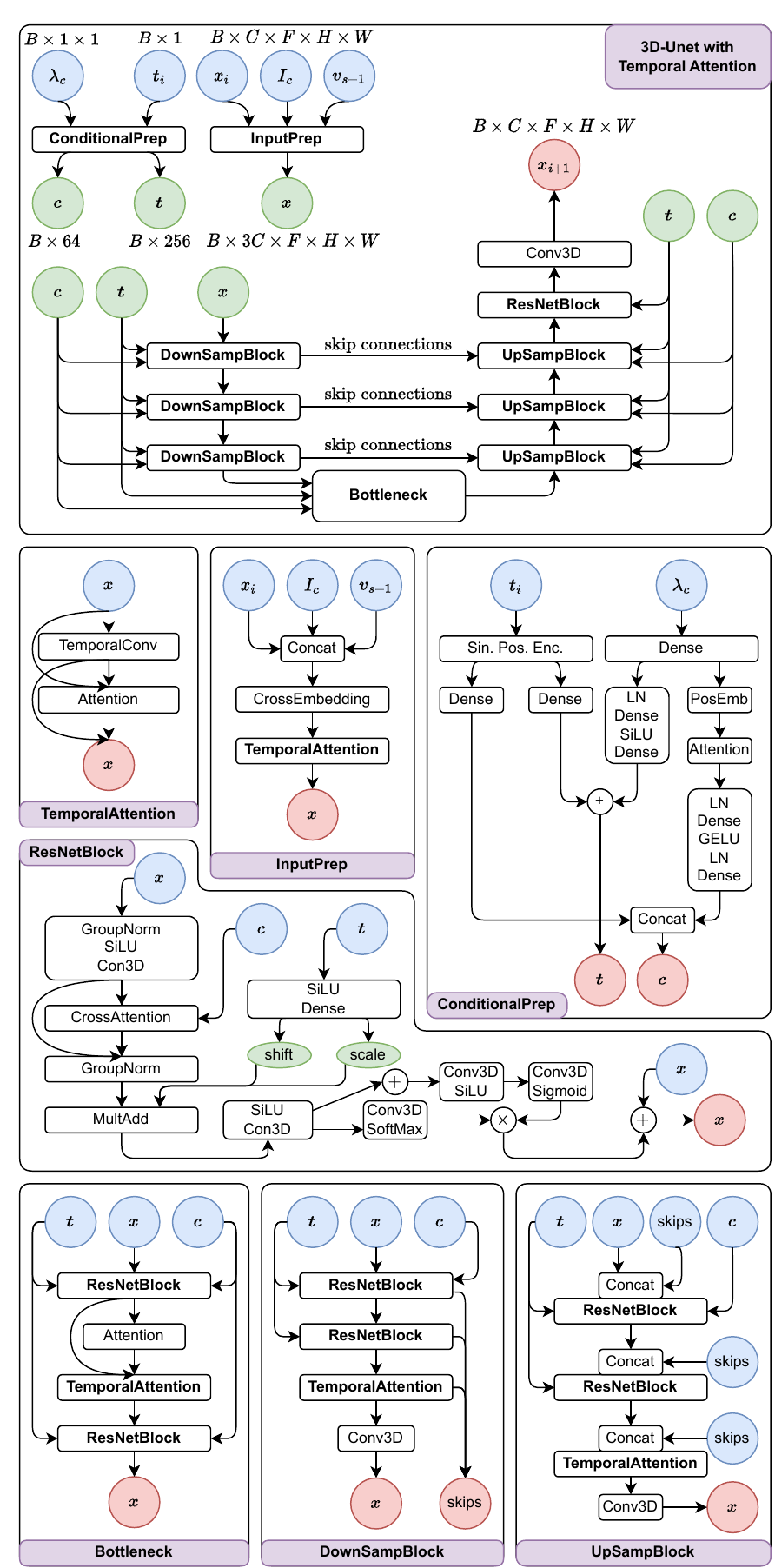}
    \caption{Model architecture. Ellipsoids and circles are variables, blue indicates an input, green an internal value and red an output. The figure is split into named groups of layers, reused layers are signalized with a bold name. We do not set values for the dimensions, as this depends on model parametrization. The top-level inputs are the noised video input $x_i$, the corresponding time conditioning $t_i$, the step-independent arbitrary conditioning $\lambda_c$, the reference frame $I_c$ and, when applicable, the previous stage output $v_{s-1}$. The final output $x_{i+1}$ is a slightly denoised version of $x_i$.}
    \label{fig:my_label}
\end{figure}

\end{document}